\documentclass[lettersize,journal]{IEEEtran}
\usepackage{amsmath,amsfonts}
\usepackage{algorithmic}
\usepackage{algorithm}
\usepackage{array}
\usepackage[caption=false,font=footnotesize,labelfont=rm,textfont=rm]{subfig}
\usepackage{textcomp}
\usepackage{stfloats}
\usepackage{url}
\usepackage{verbatim}
\usepackage{graphicx}
\usepackage{cite}
\usepackage[T1]{fontenc}
\usepackage{booktabs}
\usepackage{multirow}
\usepackage{footnote}
\usepackage{threeparttable}
\usepackage{svg}

\begin{document}

\title{YOLOMG: Vision-based Drone-to-Drone Detection with Appearance and Pixel-Level Motion Fusion}

\author{Hanqing Guo, Xiuxiu Lin, Shiyu Zhao\vspace{-2em}
\thanks{The authors are with the Department of Artificial Intelligence, Westlake University, Hangzhou, China (e-mail: guohanqing@westlake.edu.cn, zhaoshiyu@westlake.edu.cn)}
}

\maketitle

\begin{abstract}
Vision-based drone-to-drone detection has attracted increasing attention due to its importance in numerous tasks such as vision-based swarming, aerial see-and-avoid, and malicious drone detection. However, existing methods often encounter failures when the background is complex or the target is tiny. This paper proposes a novel end-to-end framework that accurately identifies small drones in complex environments using motion guidance. It starts by creating a motion difference map to capture the motion characteristics of tiny drones. Next, this motion difference map is combined with an RGB image using a bimodal fusion module, allowing for adaptive feature learning of the drone. Finally, the fused feature map is processed through an enhanced backbone and detection head based on the YOLOv5 framework to achieve accurate detection results. To validate our method, we propose a new dataset, named ARD100, which comprises 100 videos (202,467 frames) covering various challenging conditions and has the smallest average object size compared with the existing drone detection datasets. Extensive experiments on the ARD100 and NPS-Drones datasets show that our proposed detector performs exceptionally well under challenging conditions and surpasses state-of-the-art algorithms across various metrics. We publicly release the codes and ARD100 dataset at \emph{https://github.com/Irisky123/YOLOMG}.
\end{abstract}

\section{Introduction}

Vision-based drone-to-drone detection has attracted increasing attention in recent years due to its application in many tasks such as vision-based swarming\cite{tang2018vision, 2020marker}, aerial see-and-avoid\cite{2021sense-avoid}, and malicious drone detection\cite{2022detection}. This task is more difficult than general object detection because the camera itself is moving and the target drone is often obscured by complex backgrounds such as buildings and trees when viewed from an aerial angle. Additionally, when observed from a considerable distance, the target drone may appear as an extremely small object within the camera’s frame. 
\begin{figure}[!t]
	\centering
	\subfloat[Background confusion]{\includegraphics[width=0.99\linewidth]{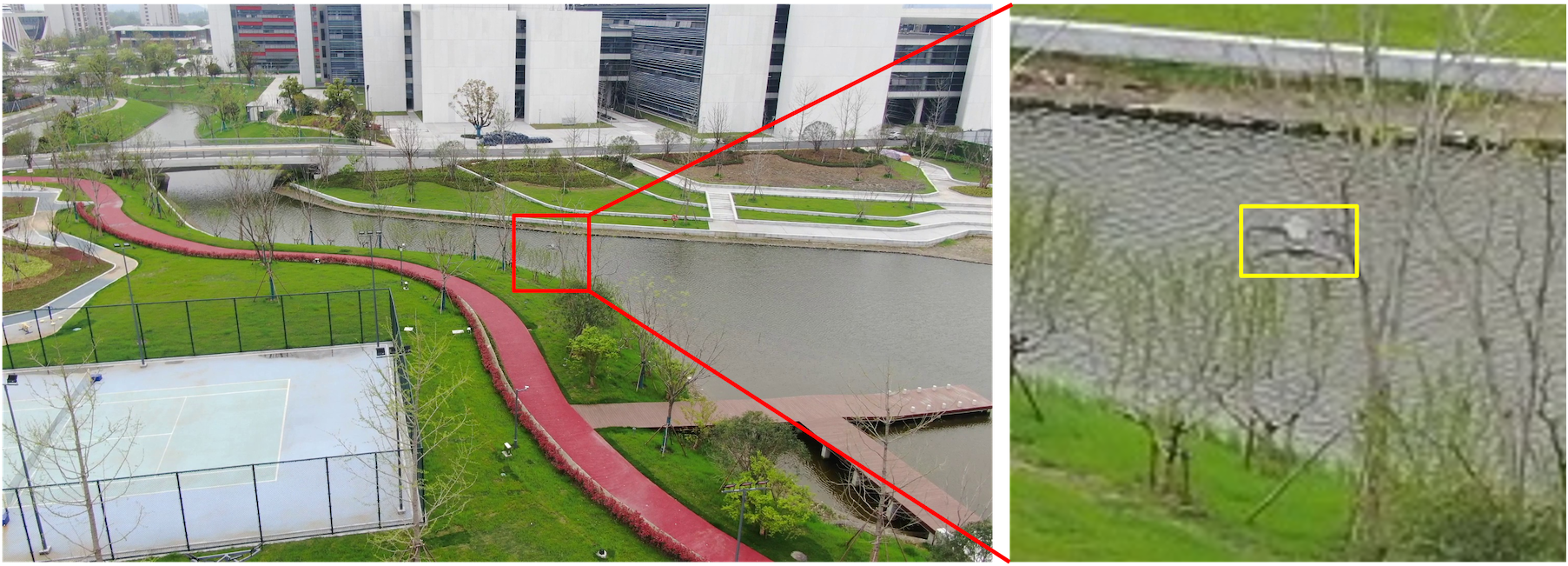}\label{fig1_complex}}\vspace{0.1mm}
	\subfloat[Tiny object]{\includegraphics[width=0.99\linewidth]{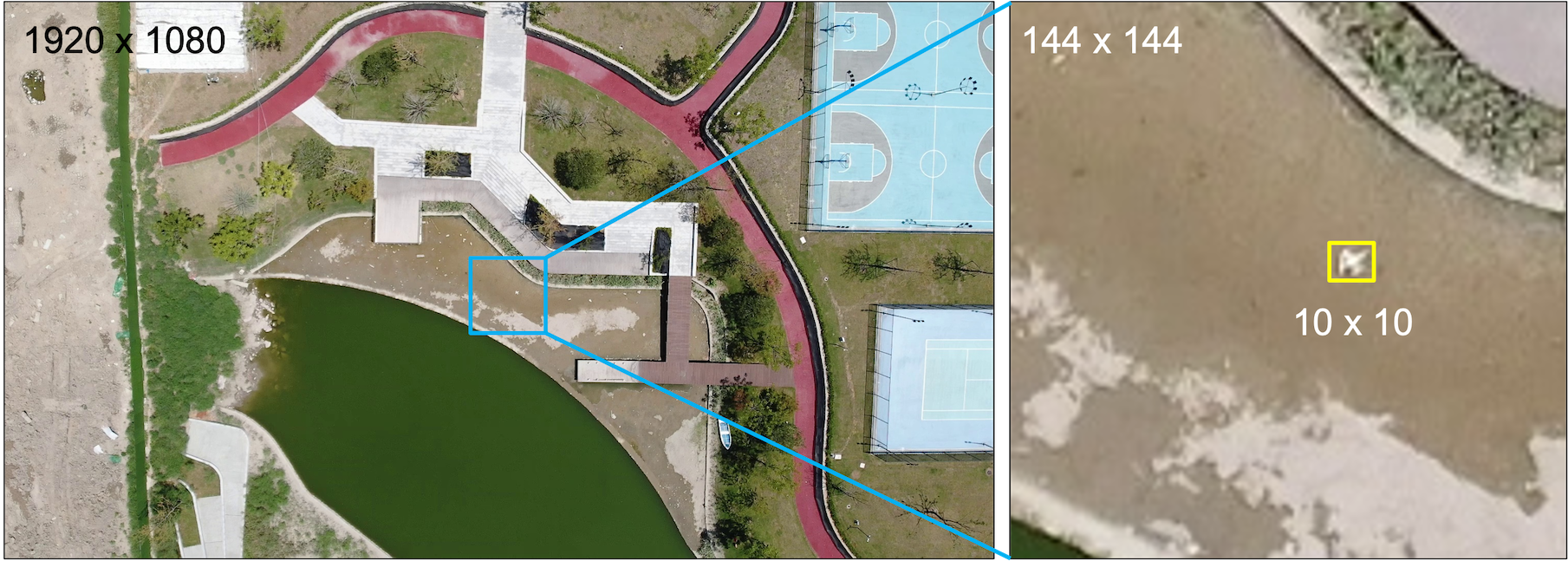}\label{fig1_tiny}}\vspace{0.1mm}
        \subfloat[Camera movement]{\includegraphics[width=0.99\linewidth]{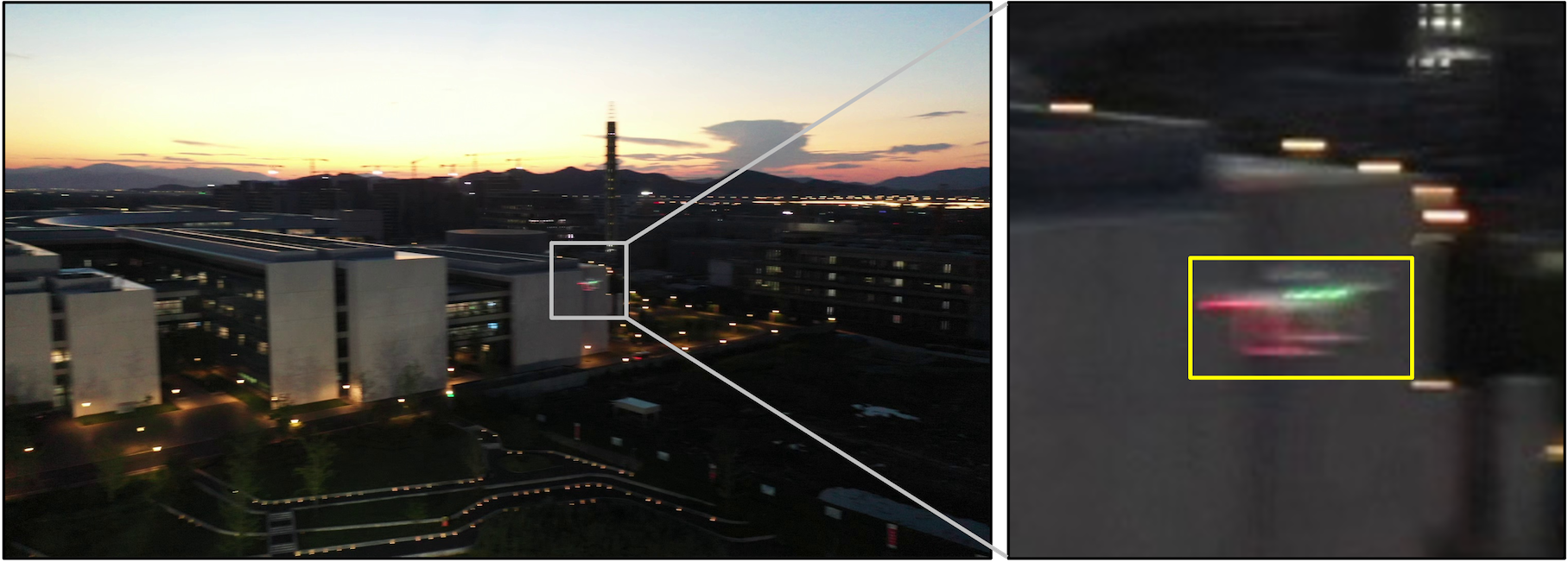}}
	\caption{Illustration of various challenges in drone-to-drone detection. (a) shows that the drone is confused with the background, (b) indicates that the objects in our ARD100 dataset are extremely small, (c) shows the motion blur caused by the camera's ego-motion. (right images are better view with 700$\%$ zoom in).}
	\label{fig_1}
\end{figure}

Some standard object detection networks such as YOLO series, R-CNN series, SSD, and DETR have been applied to drone detection \cite{2021DT-Benchmark, 2021Air, 2022Anti-UAV-DT}. These methods can work effectively in simple scenarios where the target drone is distinct and large relative to the background. However, they often encounter failures in more complex scenarios where the appearance features are unreliable. For example, as shown in Fig.~\ref{fig1_complex}, when the background scene is extremely complex, the target drone can easily get engulfed in the background scenes.
Moreover, when the target drone flies far away from the camera, it may occupy only a tiny portion of the image. For example, as shown in Fig.~\ref{fig1_tiny}, a drone seen from about 100~m away only occupies 10 × 10 pixels in an image of 1920 × 1080 pixels. Additionally, downsizing and pooling adopted by most object detection networks can further exacerbate these issues. Therefore, it is imperative to develop effective algorithms for drone detection under challenging conditions.

Recent methods have incorporated motion features or temporal information into drone detection under challenging conditions, such as background subtraction\cite{2021Fast, NPU2020, 2024efficient, GLAD, GUO2024PRL}, temporal information\cite{Xie2021SmallLT, 2023transvisdrone}, and optical flow\cite{Rozantsev2017DetectingFO, 2021Dogfight, wang2023RAFT}. While effective to some extent, these methods may not be optimal for scenarios where drones are extremely small and blend into urban backgrounds. These methods still face the following limitations. First, most of these methods have only been tested on the NPS-Drones\cite{2021Fast} and FL-Drones\cite{Rozantsev2017DetectingFO} datasets, and their effectiveness in more challenging conditions has not been thoroughly studied. Second, the motion feature of small drones is difficult to distinguish from the background, especially when the camera itself is moving. Third, existing methods lack generalization to new scenes and new types of drones, which is crucial for real-world applications. Lastly, many existing methods impose excessive computational costs, making them impractical for airborne platforms.

To address the aforementioned challenges, we propose a motion-guided object detector (YOLOMG) for extremely small drone detection. First, we introduce a motion feature enhancement module to extract pixel-level motion features of small drones. Next, we fuse the motion difference map with the RGB image using a bimodal fusion module to adaptively learn the drone's features. Finally, the fused feature map is processed through an enhanced lightweight backbone and head network based on YOLOv5 to produce the detection results.

The technical novelties are summarized as follows.

\begin{enumerate}
\item We introduce a new air-to-air drone detection dataset, named ARD100, which has the largest amount of videos and the smallest average object size (approximately \textbf{0.01$\%$} of frame size) among existing air-to-air drone detection datasets. It includes diverse challenging scenarios such as complex urban backgrounds, abrupt camera movement, low-light conditions, and tiny drones, significantly enhancing its utility for advancing drone detection and tracking research.

\item We propose combining a motion difference map, which serves as a pixel-level motion feature, with RGB images to detect extremely small objects. Experimental results on the ARD100 dataset indicate that our proposed method performs exceptionally well under challenging conditions and outperforms the best comparison algorithm by \textbf{22$\%$} in terms of average precision.

\item We present an efficient end-to-end framework for detecting extremely small drones, achieving state-of-the-art performance on the NPS-Drones dataset. Furthermore, our method exhibits high efficiency and strong generalizability, outperforming general object detectors across various metrics on the untrained Drone-vs-Bird dataset and under low-light conditions.

\end{enumerate}

\section{Related Work}\label{section_relatedWork}
\subsection{Vision-Based Drone Detection}
Vision-based drone detection has attracted increasing attention in recent years. Inspired by the success of general object detection, some researchers have applied state-of-the-art object detection networks to drone detection tasks. For instance, \cite{2021Air} introduced the Det-Fly dataset for air-to-air drone detection and evaluated eight deep learning algorithms within it. Similarly, \cite{2021DT-Benchmark} assessed four advanced deep learning algorithms across three representative drone datasets. To enhance detection accuracy, \cite{2021_Transfer_Adaptive_fusion} proposed a method combining transfer learning and adaptive fusion to boost small object detection performance. Additionally, \cite{2021PruneYOLOv4} developed thinner models for real-time small drone detection by pruning convolutional channels and YOLOv4 shortcut layers. However, these studies' results indicate that existing object detection networks struggle with complex backgrounds, motion blur, and small objects due to insufficient visual features.

Since drones are usually flying, motion features and temporal information are important cues to discriminate drone detection from other general object detection tasks. The work in \cite{2015flying, Rozantsev2017DetectingFO} is an early research in the field of detecting flying objects using temporal information. The authors first employ two CNN networks in a sliding window fashion to obtain the motion-stabilized spatial-temporal cubes, and then use a third CNN network to classify drones in each spatial-temporal cube. Additionally, the authors released the first drone-to-drone detection dataset, named FL-Drones, which has been widely used for drone detection research so far. The work in \cite{Li2016MultitargetDA, 2021Fast} creates a new UAV-to-UAV dataset named the NPS-Drones dataset and proposes a computationally efficient pipeline consisting of a moving target detector, followed by a target tracker. Subsequently, the work in \cite{2021Dogfight} proposes a two-stage segmentation-based approach to detect drones. In the first stage, frame-wise detections are obtained through channel and pixel-wise attention. Then, motion boundary, tracking, and I3D network are used to obtain the final detection results. Recently, the work in \cite{2023transvisdrone} has proposed an end-to-end method for drone-to-drone detection. The authors utilize the YOLOv5 backbone to learn object-related spatial features and the VideoSwin Transformer to learn the spatio-temporal dependencies of drone motion. The work in \cite{2024efficient} proposes the use of a feature alignment fusion module and a background subtraction module to improve the performance of drone detection. These methods could work fine when the targets occupy a relatively large part of the entire image or the background is comparatively simple. However, their performance on extremely small targets and complex backgrounds has not been well studied yet. 

\subsection{Small Object Detection} 
Small object detection (SOD) remains a challenge due to insufficient appearance and texture information. State-of-the-art methods often leverage temporal information or motion features to enhance tiny object detection. For instance, the work in \cite{2018clusternet} generates an object heatmap by stacking five grayscale frames and processing them with a two-stage CNN model using a coarse-to-fine approach. Other studies \cite{2021Dogfight, 2023CFINet} adopt two-stage frameworks, dividing frames into overlapping patches to preserve valuable appearance features of tiny objects. The work in \cite{2023temporal} presents a spatio-temporal deep learning model based on YOLOv5 that processes sequences of frames to exploit temporal context. The work in \cite{2023TinyAOD} locates potential objects using inconsistent motion cues between airborne objects and backgrounds, while the work in \cite{2024visible} introduces a self-reconstruction mechanism to enhance the weak representation of tiny objects through a difference map.
Despite their effectiveness in certain SOD scenarios, these methods have not been verified for drone-to-drone detection scenarios where the camera's ego motion is significant.

\section{Proposed Method}\label{methodology}
\begin{figure*}[!ht]
	\centering
	\includegraphics[width=0.99\linewidth]{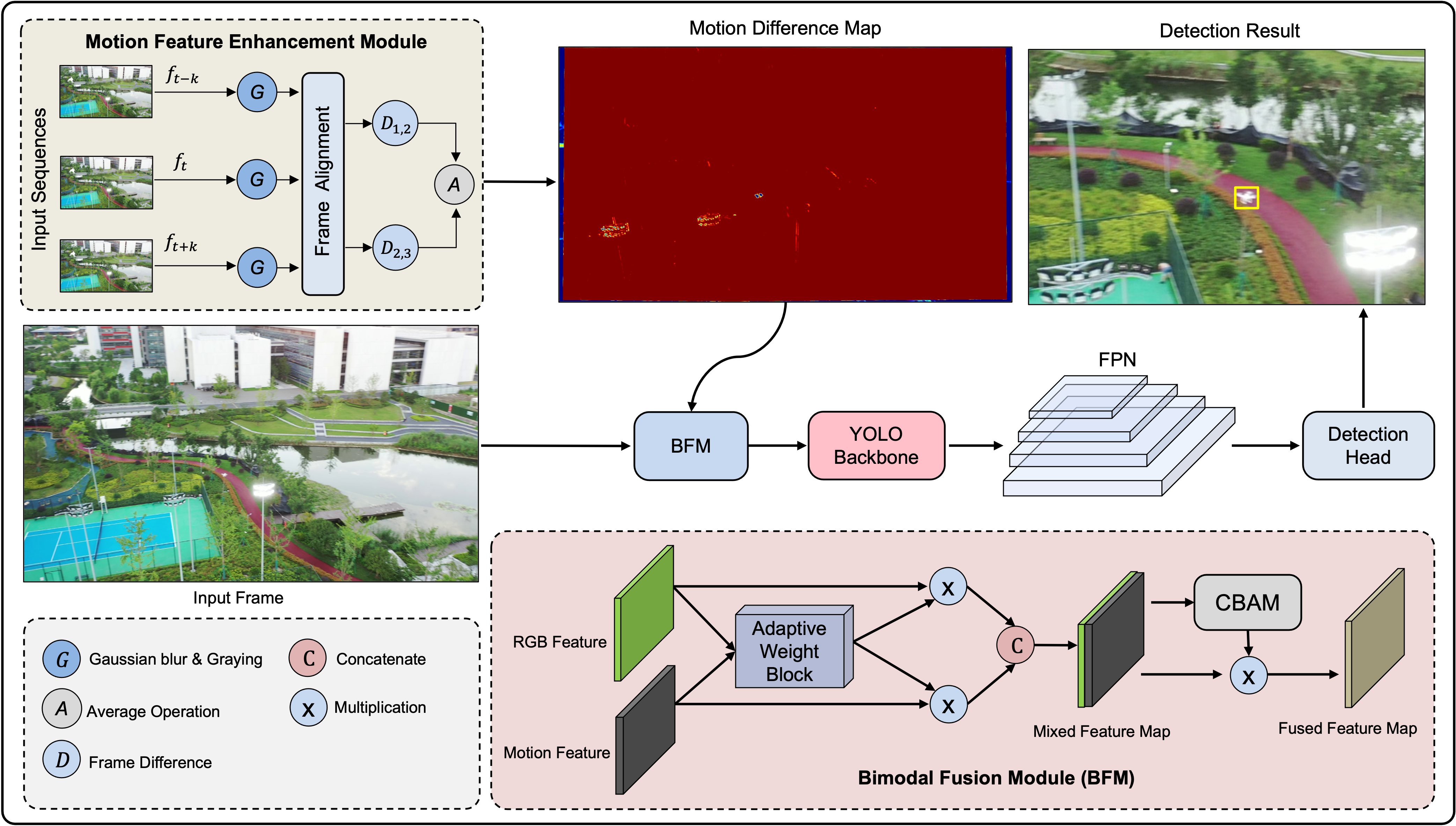}
	\caption{The overall architecture of our proposed YOLOMG algorithm. First, a motion feature enhancement module extracts the motion difference map of drones. Next, a bimodal fusion module adaptively combines the RGB and motion features. Then, the fused feature map is passed to the lightweight YOLO backbone for deep features extraction and processed through the Feature Pyramid Network (FPN) for cross-layer fusion. Finally, the feature maps are fed to the detection head to produce the detection results.}
	\label{fig_YOLOMG}
\end{figure*}

The overall architecture of our proposed YOLOMG is shown in Fig.~\ref{fig_YOLOMG}. It mainly comprises the motion feature enhancement module (MFEM) and the bimodal fusion module (BFM). The following sections detail each module.

\subsection{Motion Feature Enhancement Module} Optical flow has been adopted to extract the motion feature of drones in some studies\cite {wang2023RAFT, 2021Dogfight, 2024xsvid}. However, it may fail when the target is too small or the background is too complex, due to the lack of distinct feature points. In addition, the high computational cost of dense optical flow networks is impractical for mobile platforms. Therefore, inspired by the success of tiny object detection\cite{2024visible, 2024efficient} and video action recognition\cite{2023seeinginflowing, 2021STDN}, we propose using pixel-level difference maps to establish the motion feature of drones.

Frame difference\cite{deepfusion2021, delibacsouglu2023moving} is a simple yet effective technique for detecting moving objects by capturing pixel-level changes between images. Assuming a short acquisition time between neighboring frames, the differences are likely due to moving objects. Consistent with existing work\cite{2021Fast, 2018effective, 2023In-Flight}, we first stabilize frames using image alignment techniques and then apply a three-frame difference method to highlight moving objects.

\subsubsection{Frame Alignment}
To isolate moving pixels from the dynamic background, pixel-level frame alignment is necessary to eliminate the influence of the camera's ego-motion. In this paper, we employ 2D perspective transformation for motion compensation, as it accurately models the 2D background motion resulting from the relative movement of a 2D plane in a 3D world\cite{2021Fast, 2018effective, 2023In-Flight}.

To achieve computational efficiency and robustness in textureless regions such as the sky and grasslands, we employ grid-based key points to calculate the homography matrix. Specifically, key points are uniformly sampled in each row and column across the previous $k$ frame. These key points are then tracked using the pyramidal Lucas-Kanade (LK) algorithm\cite{PLK} to obtain their corresponding points in the current frame. Following the matching of these key points across two successive frames, the homography matrix $H$ is computed using the RANSAC method to reject outliers. The image in the previous $k$ frame $I_{t-k}$ can be aligned with the current frame $I_t$ by the perspective transformation $\hat {I}_{t-k}= \emph{H} I_{t-k}$. Here, $H$ represents the perspective transformation matrix between $I_{t-k}$ and $I_t$, $ \hat {I}_{t-k}$ denotes the motion-compensated previous $k$ frame. 

\subsubsection{Motion Difference Map Generation}
\begin{figure}[h]
	\centering
	\includegraphics[width=0.99\linewidth]{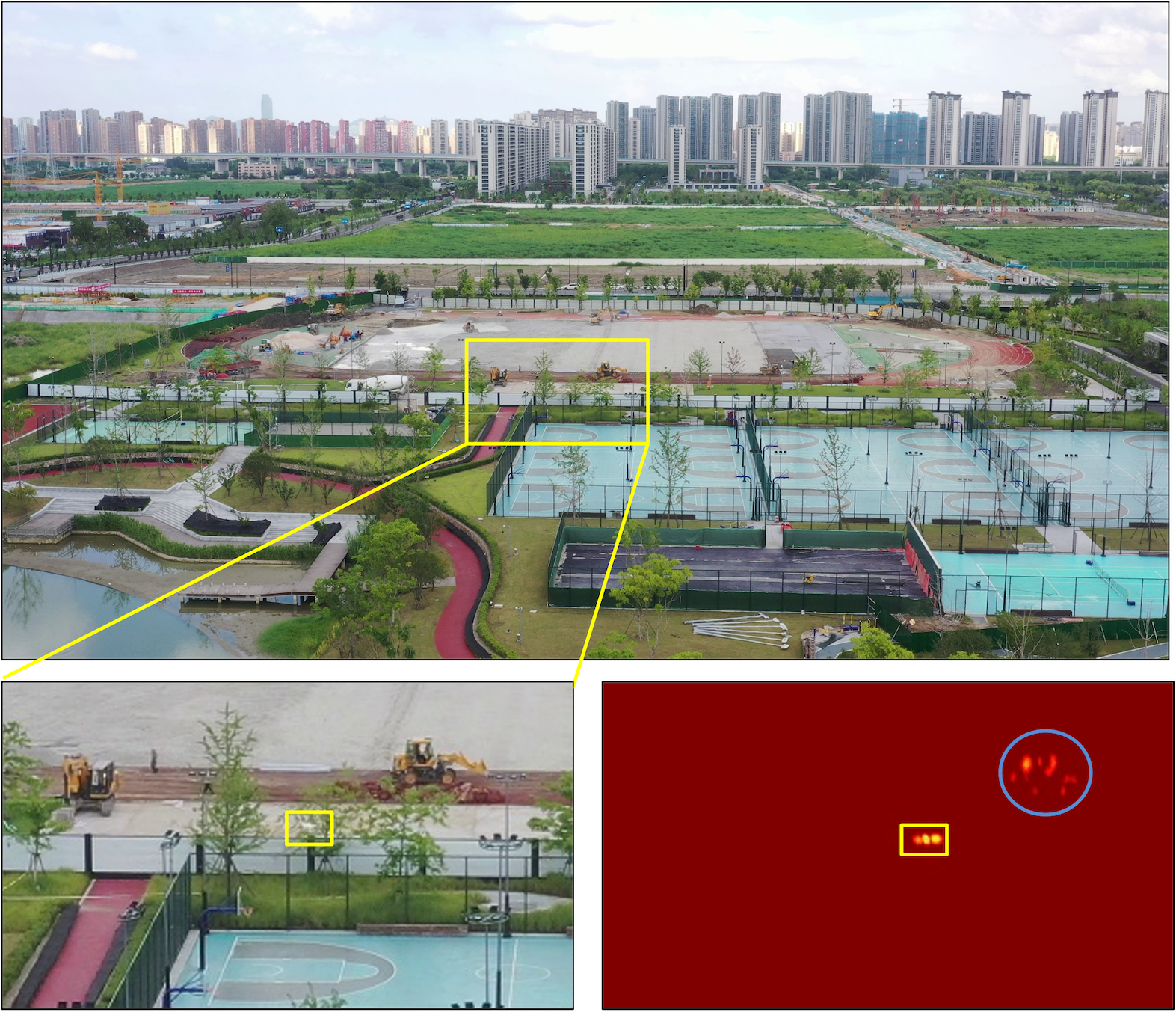}
	\caption{An example of the motion difference map. The bottom left image is the cropped RGB image, the bottom right image is the cropped motion difference map (bottom images are better view with 300$\%$ zoom in). The yellow boxes enclose the target drone. The blue circle enclose the interruptions.}
	\label{fig_motion_map}
\end{figure}

After we have obtained the motion-compensated previous $k$ frame, the moving areas are highlighted by computing the absolute differences between the current frame and the motion-compensated previous $k$ frame. To enhance the motion features of tiny moving objects while reducing noise, an improved three-frame difference method is employed, as detailed below:
\begin{equation}
	E_{t} = (|I_t - \hat {I}_{t-k}| + |I_t - \hat {I}_{t+k}|)/2,
\end{equation}
where $I_{t-k}$, $I_t$ and $I_{t+k}$ are the gray value of the previous $k$ frame, current frame, and the next $ k$ frame, $ \hat {I}_{t-k}$ and $ \hat {I}_{t+k}$ are the gray value of the motion-compensated previous $k$ frame and next $k$ frame. We first compute the frame difference between two adjacent frames. Subsequently, an average operation is applied to two adjacent frame differences to enhance the motion features of moving objects. While frame difference can segment pixels belonging to moving objects, it also introduces noise, small holes, and disconnected blobs. To address these issues, morphological open and close operations are iteratively applied to remove isolated pixels and fill the holes. An example of the motion difference map is shown in Fig.~\ref{fig_motion_map}, where it is evident that our proposed method can effectively extract the motion features of extremely small drones against complex backgrounds.

\subsection{Bimodal Fusion Module}
\subsubsection{Fusion Strategy}
The motion difference map provides positional guidance for the target but is susceptible to interruptions such as moving cars, pedestrians, flying birds, and image alignment errors (as shown in the blue circle of Fig.~\ref{fig_motion_map}). Although these interruptions resemble drones in motion difference maps, they exhibit distinct appearance characteristics in RGB images. Thus, we can leverage the appearance features of drones in RGB images to differentiate them from interruptions. Inspired by multi-modal fusion object detection\cite{2023SLBAF, 2023superyolo, deepfusion2021, 2023temporal_frame_diff}, we treat the motion difference map and the RGB image as two separate modalities of drones. We adopt a feature-level fusion strategy to integrate the information from these dual data sources while maintaining a lightweight model.
Initially, an RGB image and a motion difference map are combined using a bimodal fusion module. The resulting fused feature map is then passed to the backbone for downsampling and processed through the Feature Pyramid Network (FPN) structure for cross-layer fusion. Finally, the feature map is sent to the detection head, which generates the final detection results.

\begin{figure}[h] 
	\centering
	{\includegraphics[width=0.99\linewidth]{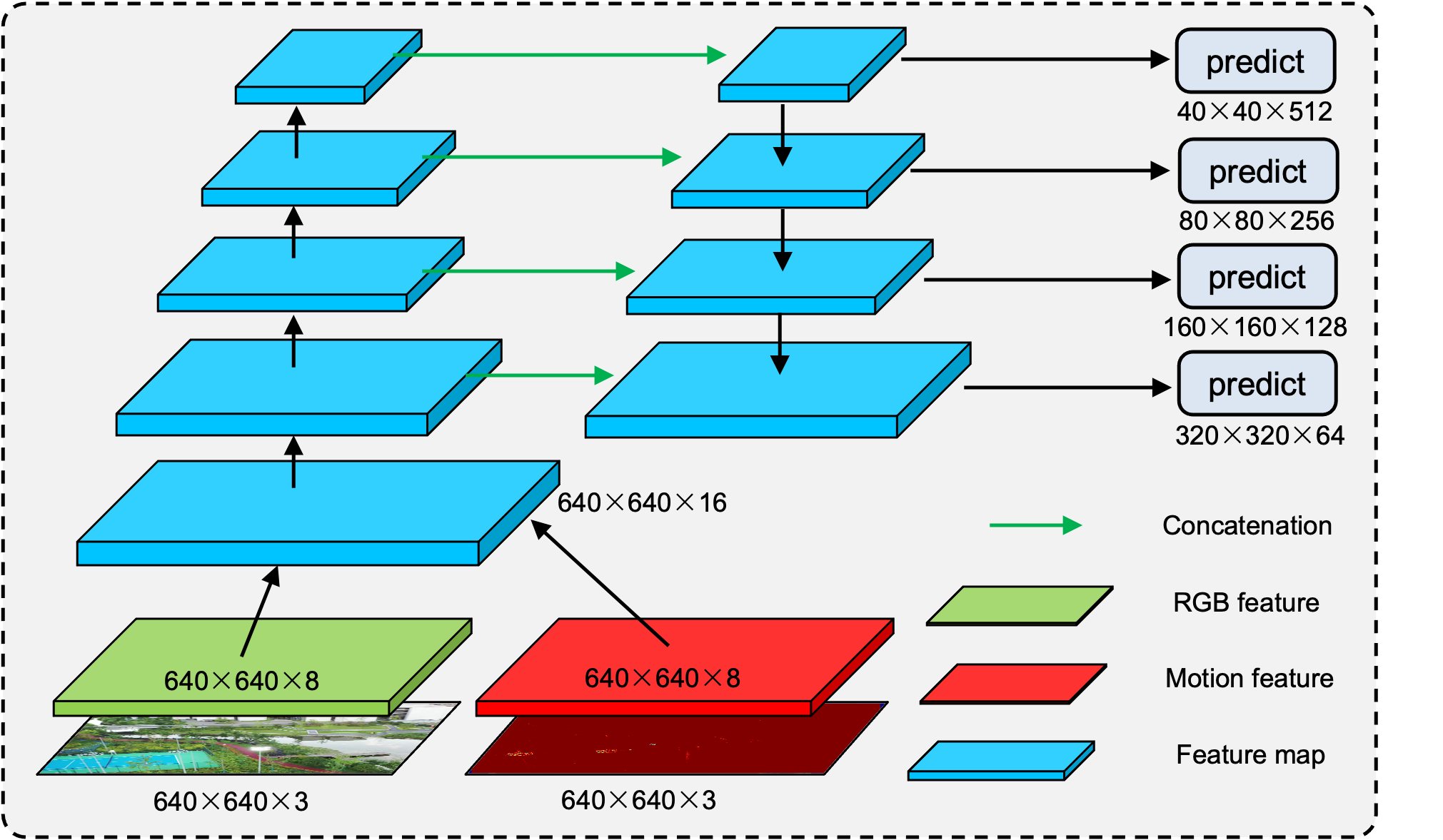}}
	\caption{The structure of our proposed YOLOMG network.}
	\label{fig_structure}
\end{figure}

\subsubsection{Network Structure}
Our method focuses on small drone detection, with the network structure depicted in Fig.~\ref{fig_structure}. The YOLOMG network comprises 35 layers and employs 16× down-sampling. Our backbone network is an extended version of the YOLOv5s backbone, which has been tailored to enhance computational speed by reducing the number of channels. To improve small target detection, we incorporate a small object detection layer by up-sampling the last layer's feature map in the neck network, combining it with the backbone's corresponding feature map, and feeding them into the detection head. This results in four detection heads for prediction. Additionally, the Feature Pyramid Network (FPN) structure combines deep and shallow features during up-sampling to create a robust, high-resolution semantic feature map. Compared to the original YOLOv5s network, our design is more lightweight and specifically optimized for small object detection.

\subsubsection{Bimodal Adaptive Fusion}
During the feed-forward process of a convolutional neural network, all features receive equal attention, which means that they are treated with the same weight. In the case of a bimodal input network, there can be situations where one input is of high quality while the other is of low quality. For example, an RGB image may not provide sufficient discriminative information for drones when the background is too complex. However, clear motion information can still be obtained from the motion difference map. 

To address these challenges, we adopt an adaptive fusion strategy similar to the approaches discussed in \cite{2023SLBAF, 2023superyolo}. Our strategy consists of two components: an adaptive weight block and a channel and spatial attention block, as illustrated in the lower right corner of Fig.~\ref{fig_YOLOMG}. First, the RGB feature map and the motion feature map generate initial weights and a mixed feature map through the adaptive weight block. Next, this mixed feature map is processed by the CBAM (Convolutional Block Attention Module) to obtain channel weights and produce the final fused feature map. The use of CBAM allows us to assign more weight to higher quality input feature maps.

\section{Experiments}\label{experiment}

\subsection{Datasets}\label{dataset}
To evaluate the performance of YOLOMG, we tested it on our proposed ARD100 dataset and the publicly available NPS-Drones dataset. 

\begin{figure*}[!t]
	\centering
        {\includegraphics[width=0.98\linewidth]{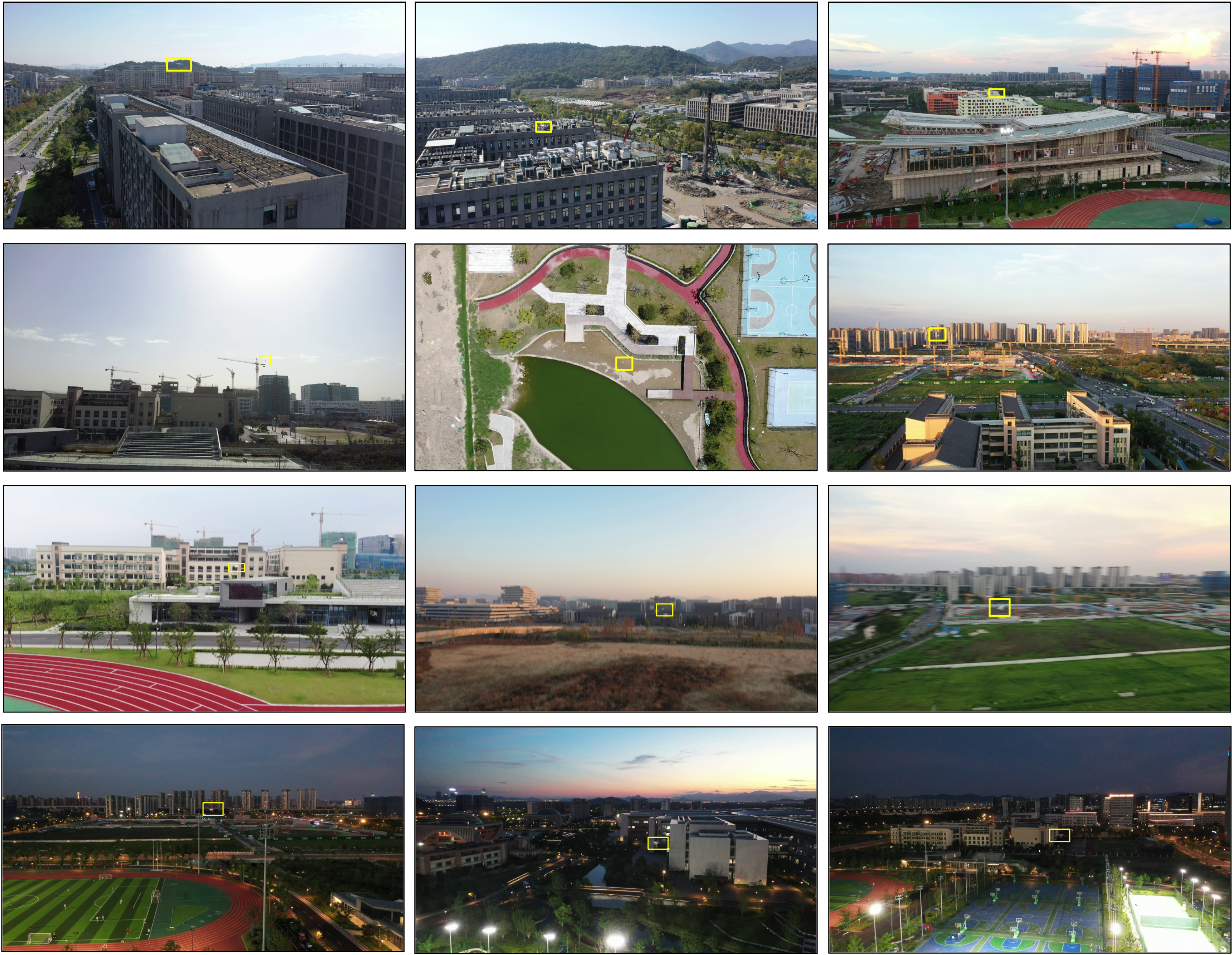}}
	\caption{Some representative images in the ARD100 dataset. The first row is the \textbf{complex backgrounds}, the second row is the \textbf{tiny objects} (usually smaller than 12$\times$12 pixels), the third row demonstrates the severe motion blur caused by \textbf{abrupt camera movement}, and the fourth row shows the drones under \textbf{low-light conditions}.}
	\label{fig_dataset}
\end{figure*}

\textbf{ARD100 dataset }
This dataset comprises 100 video sequences with a total of 202,467 frames. All videos were captured using DJI Mavic2 or DJI M300 cameras at low and medium altitudes. As depicted in Fig.~\ref{fig_dataset}, the dataset encompasses numerous real-world challenges, including complex backgrounds, low and strong light conditions, abrupt camera movement, fast-moving drones, and extremely small drones. Each video is recorded at a frame rate of 30 FPS with a resolution of 1920 × 1080.

A comprehensive comparison between the ARD100 dataset and other significant drone detection datasets is presented in Table~\ref{tab_dataset}. It is evident that objects smaller than 12 × 12 pixels constitute the largest proportion in the ARD100 dataset. To the best of our knowledge, this dataset features the largest number of videos and the smallest average object size compared to existing drone-to-drone detection datasets. We allocated 65 videos for training and 35 videos for testing. Notably, the test set includes scenes absent from the training set, as well as videos captured under low- and strong-light conditions, to verify the model's generalization ability and robustness.

\begin{table*}[h]
	\centering
        \caption{Comparison of various datasets for drone detection.}
	\label{tab_dataset}
			\begin{tabular}{ccccccc}
				\toprule 
                    \multicolumn{1}{c}{\multirow{2}{*}{Dataset}}&\multirow{2}{*}{Frames}&\multirow{2}{*}{Videos}&\multicolumn{3}{c}{Area Size (Pixels)} & \multirow{2}{*}{Average Area / Frame Size}\\ \cmidrule{4-6}
                    \multicolumn{1}{c}{} & & & 0 \textasciitilde 12$^{2}$ & 12$^{2}$ \textasciitilde 20$^{2}$ & 20$^{2}$ \textasciitilde 32$^{2}$& \\
				\midrule 
				NPS-Drones\cite{2021Fast} & 70,250 & 50 & 35.91$\%$ & 59.85$\%$& 4.10$\%$ & 0.05$\%$ \\
				FL-Drones\cite{Rozantsev2017DetectingFO}  & 38,948  & 20 & 20.40$\%$ & 50.65$\%$ & 20.50$\%$ & 0.07$\%$ \\
				Drone-vs-Bird\cite{drone-vs-bird} & 104,760 & 77 &19.93$\%$&30.09$\%$&22.82$\%$& 0.1$\%$  \\
				ARD100 & 202,467 & 100 & 42.18$\%$ & 37.55$\%$ & 16.30$\%$ & \textbf{0.01$\%$}  \\
				\bottomrule 
			\end{tabular}
	\end{table*}
 
\textbf{NPS-Drones dataset\cite{2021Fast}}
The NPS-Drones dataset comprises 50 videos of a custom delta-wing aircraft, totaling 70,250 frames. The videos were captured using a GoPro 3 camera mounted on a custom delta-wing airframe, with resolutions of either 1920 × 1280 or 1280 × 960. The objects in this dataset are primarily small drones, with sizes ranging from 10 × 8 to 65 × 21 pixels, and an average object size of 0.05$\%$ of the entire image. We utilized the clean version annotations released by \cite{2021Dogfight}. Following the train/val/test split proposed by \cite{2021Dogfight}, we allocated the first 40 videos for training and validation, and the remaining 10 videos for testing.

\subsection{Metrics and Implementation Details}
\subsubsection{Metrics} Following the related works \cite{2021Dogfight, 2023transvisdrone}, performance evaluation is based on commonly used metrics such as precision, recall, and average precision (AP) in this experiment. We set the IOU threshold between predictions and ground truth at 0.5; therefore, detected targets that match ground truth with IOU $\geq$ 0.5 are counted as true positives. In addition, we employ frames per second (FPS) tested on a single GPU as the efficiency metric.

\subsubsection{Implementation Details}
Our experiments are implemented on NVIDIA Geforce RTX 2080Ti GPUs. For the training of the YOLOMG algorithm, we use the Adam optimizer with a momentum of 0.937 and an initial learning rate of 0.01. We trained the model for 100 epochs with a batch size of 8, and start from the pre-trained YOLOv5s weight on MS COCO dataset.

\subsection{Comparison with State-of-the-art Methods}\label{comparison}
We compare our proposed YOLOMG algorithm with the state-of-the-art ones: one-stage object detector\cite{yolov9}, two-stage object detector\cite{fastercnn, cai2018cascade}, video object detector\cite{mega, 2023lstfe}, and small object detector \cite{2023CFINet, 2021Dogfight, 2023transvisdrone}. To retain the appearance feature of the small target as much as possible, most comparing algorithms use an input image size of 1280 × 1280 for training and testing. In particular, the training images of CFINet and Dogfight are first divided into 3 × 2 and 3 × 3 overlapping patches respectively. All compared methods are implemented based on official codes and are fine-tuned using the pre-trained weights available with public codes.

The quantitative comparison of YOLOMG with other methods on the ARD100 dataset is presented in Table~\ref{tab_sota2}. The experimental results demonstrate that our proposed algorithm outperforms existing methods across various metrics. Specifically, our method outperforms the best prior method by 21$\%$ absolute on AP metric. Notably, even with an input size of 640 × 640, the detection performance of the proposed algorithm also exceeds all comparison methods using an input size of 1280 × 1280. Furthermore, the inference speed of our algorithm at 640 × 640 input size is comparable to YOLOv5, indicating its potential for deployment on mobile platforms. The comparison with the baseline YOLOv5s model shows that the incorporation of motion features effectively enhances the model's ability to detect small targets. It is worth noting that the proposed YOLOMG network is plug-and-play and can be further optimized with new versions of the YOLO series detectors.

\begin{figure*}[t]
        \centering
	{\includegraphics[width=0.99\linewidth]{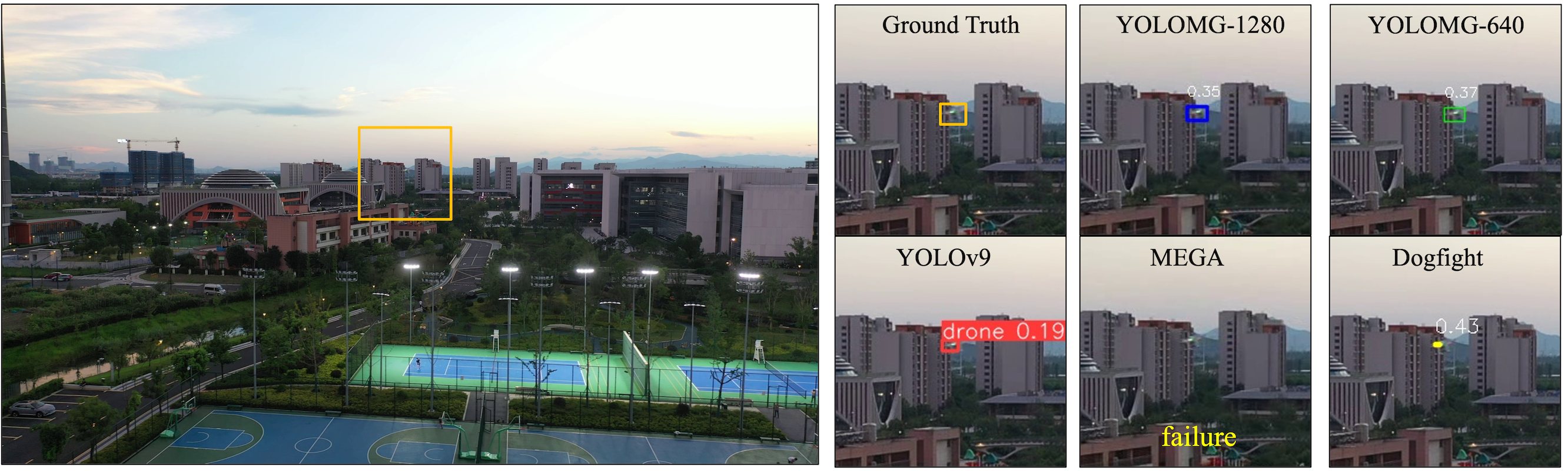}}\vspace{1mm}
        {\includegraphics[width=0.99\linewidth]{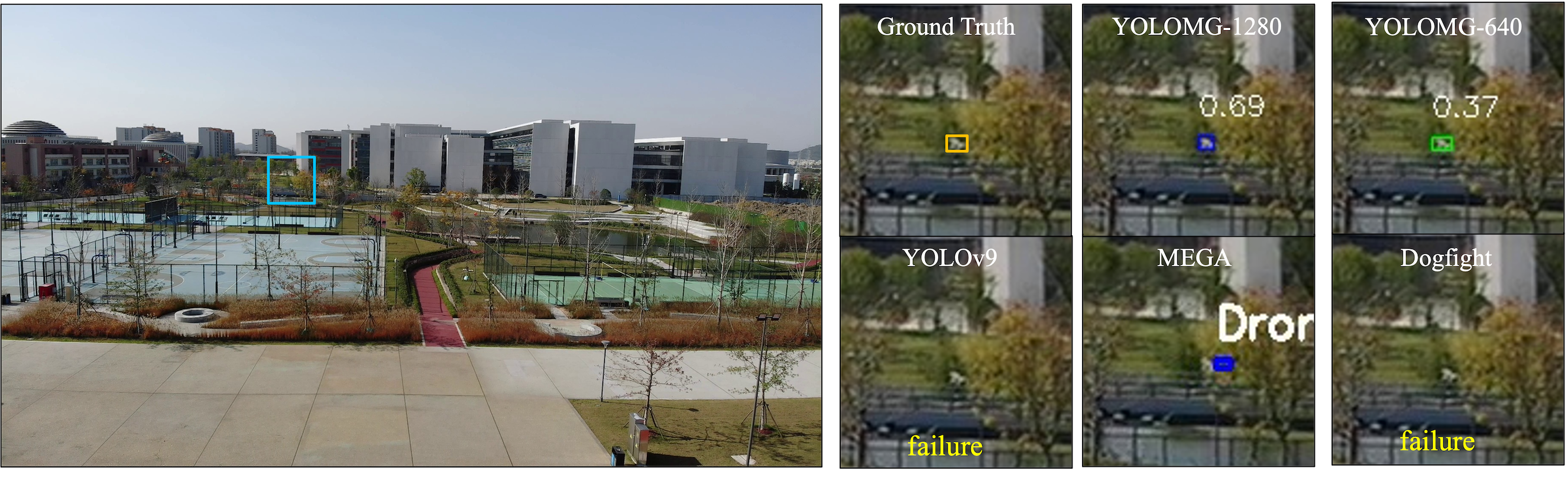}}
	\caption{Visual comparison of our YOLOMG with other compared methods. YOLOMG achieves the most accurate predictions matching the ground-truth bounding boxes and significantly outperforms other methods in the complex background and extremely small target scenarios. YOLOMG-640 indicates using a 640 × 640 input size for inference and YOLOMG-1280 indicates using a 1280 × 1280 input size for inference. The top right images are with 250$\%$ zoom in, the bottom right images are with 500$\%$ zoom in. Since the frame size is too large compared to the target's size, we just show the results on the cropped regions for a better view.}
	\label{fig_compare}
\end{figure*}

A visual comparison of YOLOMG with other methods is presented in Fig.~\ref{fig_compare}. The results demonstrate that our proposed method can effectively detect extremely small drones and drones under complex backgrounds. In contrast, the compared methods either fail to detect the target drone or produce inaccurate bounding boxes under these challenging conditions. See more test results in the supplementary videos. 

We have also evaluated YOLOMG on the NPS-Drones dataset. Some of the experimental results come from \cite{2021Dogfight} and \cite{2023transvisdrone}. As reported in Table~\ref{tab_sota2}, our proposed algorithm achieves the same highest average precision as the previous best method.

\begin{table}[!h]
	\centering
        \caption{Compared with state-of-the-art methods on the ARD100 and NPS-Drones datasets. The best results are highlighted in bold. Evaluation metrics include AP (precision) and FPS (speed).}
	\label{tab_sota2}
		\begin{tabular}{lccc}
			\toprule 
			Methods & NPS-Drones & ARD100 & FPS\\
			\midrule
                YOLOv5 & 0.93 & 0.53 & 133\\
			YOLOv8 & 0.93 & 0.53 & 103\\
                YOLOv9\cite{yolov9} & 0.91 & 0.64 & 62\\
                YOLOv11 & 0.92 & 0.55 & \textbf{233}\\
                YOLOv5-tph\cite{TPH-YOLOv5} & 0.92 & 0.51 & 120\\
			MEGA\cite{mega}& 0.83 & 0.23 & 8\\
                LSFTE\cite{2023lstfe} & - & 0.17 & 3\\
                Faster R-CNN\cite{fastercnn} & - & 0.20 & 8\\
                Cascade R-CNN\cite{cai2018cascade} & - & 0.24 & 7\\
                CFINet\cite{2023CFINet} & 0.90 & 0.63 & 32\\
                Dogfight\cite{2021Dogfight} & 0.89 & 0.50 & 1\\ 
                TransVisDrone\cite{2023transvisdrone} & \textbf{0.95} & 0.15 & 5 \\
                YOLOMG-640 & 0.92 & 0.78 & 133\\
                YOLOMG-1280 & \textbf{0.95} & \textbf{0.85} & 35\\
			\bottomrule 
		\end{tabular}
\end{table}

\subsection{Ablation studies}
\textbf{Difference map:} To test the impact of different motion difference maps, we conduct experiments with different kinds of difference maps and frame steps. Experiments are performed on the ARD100 dataset with 640 × 640 resolution. The test results are shown in Table~\ref{tab_ablation}. We can see that the motion difference map generated by the three-frame difference method is better than the two-frame difference method. In addition, a two-frame interval achieves better performance than a one-frame. However, more frames and larger steps will reduce the precision metric. This might be because the larger interval wound enlarges the motion feature of drones but may also introduce more noise. 
\begin{table}[h]
	\centering
        \caption{Ablation studies on the ARD100 dataset.}
	\label{tab_ablation}
			\begin{tabular}{lccc}
				\toprule 
				Method & Precision & Recall & AP\\
				\midrule 
				2-frame difference, k = 1 & 0.79 & 0.67 & 0.73\\
				2-frame difference, k = 2 & 0.85 & 0.69 & 0.77\\
				3-frame difference, k = 1 & 0.83 & 0.68 & 0.75\\
				3-frame difference, k = 2 & 0.83 & 0.71 & 0.78 \\
                    RGB + RGB & 0.53 & 0.31 & 0.33 \\
                    w/o small detection layer & 0.82 & 0.70 & 0.76\\
				\bottomrule 
			\end{tabular}
\end{table}

We have also conducted experiments with different kinds of difference map and frame steps on the NPS-Drones dataset with 1280 resolution. As shown in Fig.~\ref{tab_ablation_nps}, the motion difference map generated by three-frame differences is better than the two-frame differences, and a two-frame interval achieves better performance than a one-frame and a three-frame interval. This might be because the larger interval wound enlarges the motion feature of drones but may also introduce more noise. 

\begin{table}[h]
	\centering
        \caption{Ablation study of different motion difference maps on the NPS-Drones dataset}
	\label{tab_ablation_nps}
		\begin{tabular}{lccc}
				\toprule 
				Method & Precision & Recall & AP\\
				\midrule 
				3-frame difference, step = 1 & 0.93 & 0.86 & 0.92\\
				3-frame difference, step = 2 & 0.93 & 0.90 & 0.95\\
				3-frame difference, step = 3 & 0.91 & 0.90 & 0.94\\
				2-frame difference, step = 2 & 0.91 & 0.85 & 0.90 \\
				\bottomrule 
		\end{tabular}
\end{table}

To verify the effectiveness of the motion difference map, we have also tested different network inputs. We see that with a motion difference map, our algorithm performs much better than with two RGB frames. It demonstrates that the introduction of motion difference map can effectively improve the detection ability towards small targets.

\textbf{Small object detection layer:} To verify the effectiveness of the small object detection layer, we conduct experiments without the small object detection layer. As the test results in the last row of Table~\ref{tab_ablation} show, the small object detection layer effectively improves the baseline method by 2$\%$ absolute on AP metric on ARD100 dataset.

\textbf{Generalization test:} Model generalization is an important problem in object detection domain, it significantly influences the algorithm's robustness in real-world tasks. To test the robustness of our proposed method, we use well-trained weights on the ARD100 dataset to test on the videos from Drone-vs-Bird\cite{drone-vs-bird} dataset which contains different kinds of drones and different backgrounds. As shown in Table~\ref{tab_generalization}, our proposed YOLOMG algorithm performs much better than general object detectors in new environments and new drones, especially on recall metric. 

\begin{figure}[h]
	\centering
        {\includegraphics[width=0.99\linewidth]{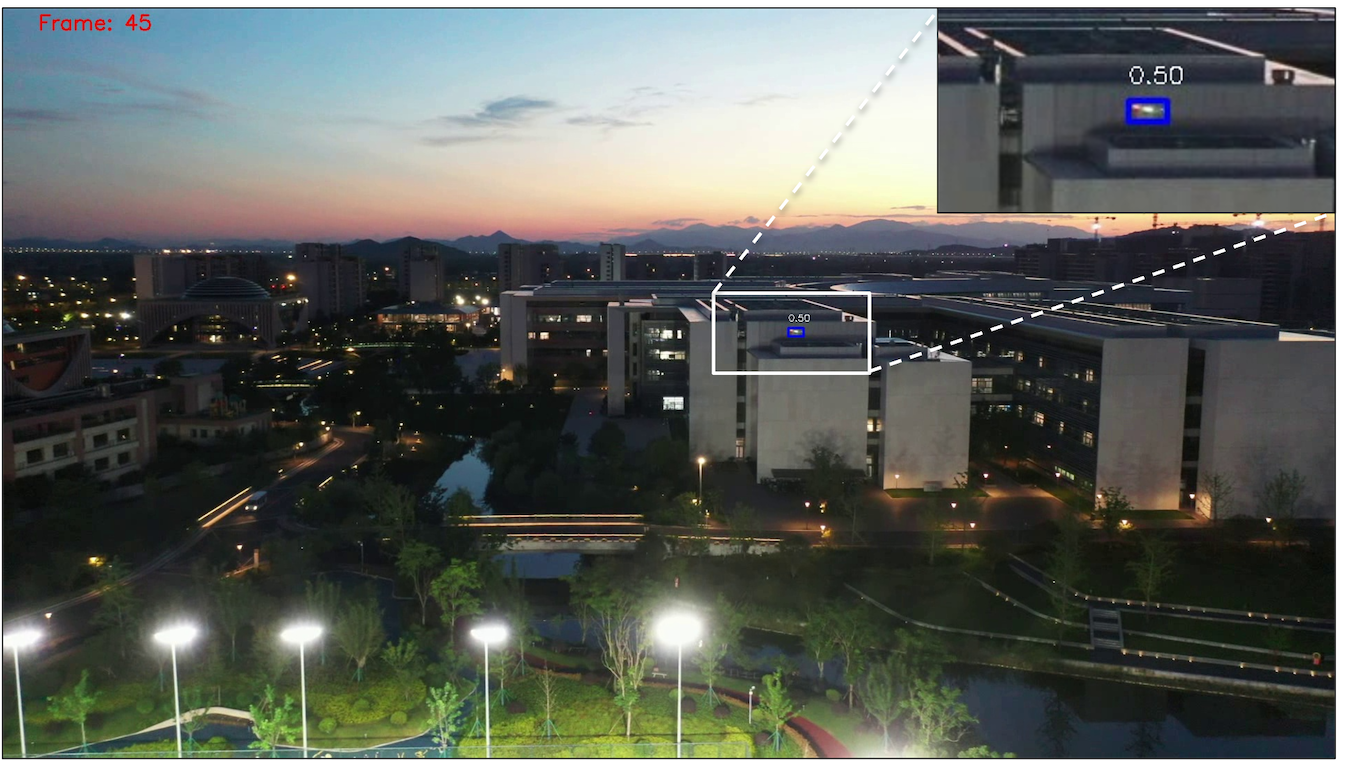}}
	\caption{Generalization test under low-light condition.}
	\label{fig_general}
\end{figure}

In addition, we have also tested the generalizability under low-light conditions. Specifically, we trained the algorithms using the ARD100 dataset, which excludes low-light scenarios, and subsequently evaluated the detection accuracy on data collected at night. As illustrated in Table~\ref{tab_generalization}, our proposed algorithm significantly outperforms the general object detectors across all evaluation metrics, thereby demonstrating superior generalization ability. An exemplary result is depicted in Fig.~\ref{fig_general}.

\begin{table}[h]
	\centering
        \caption{Model generalization test on untrained dataset and low-light conditions.}
	\label{tab_generalization}
			\begin{tabular}{l|cccc}
				\toprule 
				Scenario & Method & Precision & Recall & AP\\
				\midrule 
                \multicolumn{1}{l|}{\multirow{3}{*}{Drone-vs-Bird}}& YOLOMG & 0.50 & \textbf{0.47} & \textbf{0.41}\\
                \multicolumn{1}{l|}{}  & YOLOv9 & \textbf{0.61} & 0.28 & 0.33\\
                \multicolumn{1}{l|}{}  & YOLOv5 & 0.32 & 0.18 & 0.19\\
                    \midrule
                \multicolumn{1}{l|}{\multirow{3}{*}{Low-light}}& YOLOMG & \textbf{0.90} & \textbf{0.74} & \textbf{0.84} \\
                \multicolumn{1}{l|}{}  & YOLOv9 & 0.75 & 0.54 & 0.64 \\
                \multicolumn{1}{l|}{}  & YOLOv5 & 0.76 & 0.48 & 0.62 \\
				\bottomrule 
			\end{tabular}
\end{table}

\begin{figure}[h]
	\centering
	\subfloat[Hovering drones are occasionally ignored by our approach.]{\includegraphics[width=0.99\linewidth]{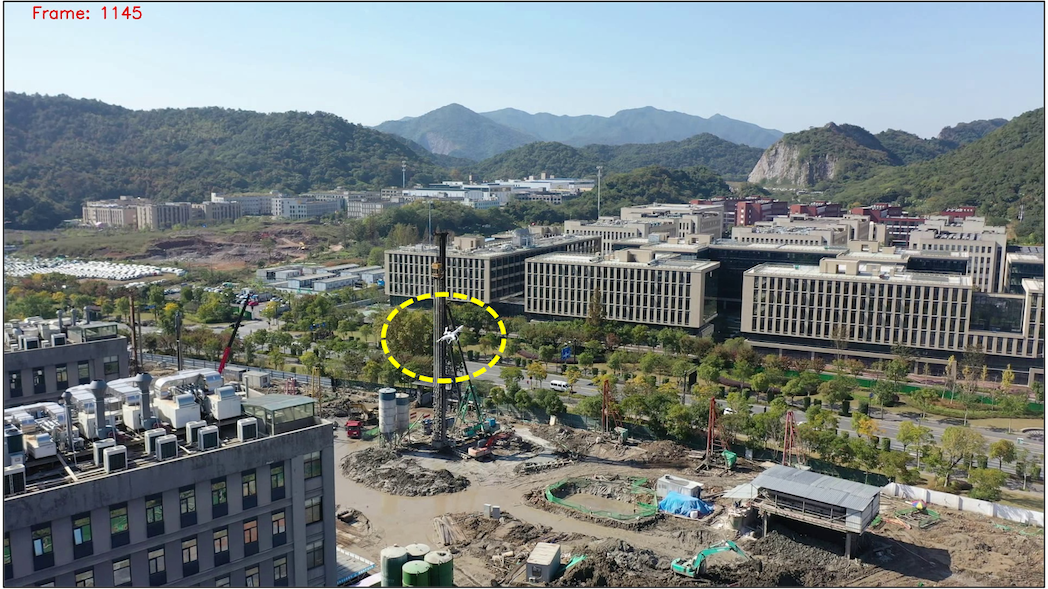}}
        \vspace{0.1mm}
	\subfloat[False detection on small objects that look similar to distant drones.]{\includegraphics[width=0.99\linewidth]{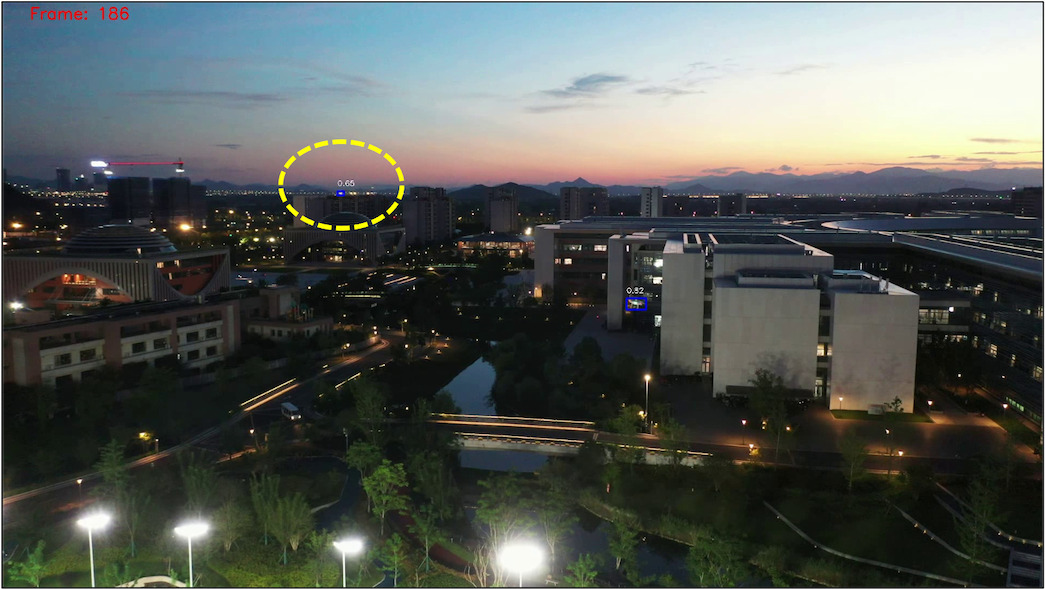}}
	\caption{Examples of the failure case.}
	\label{fig_failure}
\end{figure}

\textbf{Failure case:} Since the proposed approach tries to detect drones with a motion feature guidance, hovering and slow motion drones are occasionally ignored. In addition, since extremely small targets have very obscured appearance features, objects that look similar to distant drones are sometimes falsely detected as drones. Fig.~\ref{fig_failure} shows the failure cases of our approach.

\subsection{Discussion}
Our method focuses on detecting extremely small drones against complex backgrounds using motion features. Experimental results demonstrate that our proposed method outperforms state-of-the-art methods on various metrics and generalization ability. This is attributed to two main factors: First, YOLOMG uses a motion difference map to capture drone-specific features, which significantly enhances available information for object detection when targets are tiny or backgrounds are complex. Appearance-based detectors struggle under such conditions due to their reliance on appearance features alone. Existing methods that incorporate motion features, such as those in \cite{2021Dogfight, wang2023RAFT}, also face difficulties because their extracted motion or temporal information is insufficient for extremely small targets. Second, unlike appearance features that are class-specific, the motion difference map acts as a domain-invariant feature across various drones and environments, thus offering better generalization than general object detectors.

Despite its advantages, our approach has limitations that need to be resolved in future work. First, reliance on motion features means that stationary or slow-moving drones are occasionally missed. A network capable of simultaneously learning stationary and moving drones is needed. Second, incorporating a motion difference map increases computational costs. Therefore, it is essential to develop a more efficient network to utilize motion information.

\section{Conclusion}\label{conclusion}
This paper presents an end-to-end framework for detecting extremely small drones. We generate a motion difference map to capture motion features of small objects and fuse it with RGB images via a bimodal adaptive fusion network. To evaluate the effectiveness of our method, we introduce the ARD100 dataset, which features complex backgrounds, abrupt camera movement, low-light conditions, and tiny drones. Experiments on the ARD100 and NPS-Drones datasets show that our approach effectively detects small drones and surpasses state-of-the-art methods.

\bibliography{main} 
\bibliographystyle{ieeetr}

\end{document}